\documentclass[10pt,journal]{IEEEtran}
\IEEEoverridecommandlockouts
\usepackage{ifpdf}
\usepackage{cite}
\usepackage[hyphens]{url}
\usepackage{graphicx}
\usepackage[cmex10]{amsmath}
\usepackage{mathtools}%
\usepackage{algorithmic}
\usepackage{array}
\usepackage{stfloats}

\begin{document}
\title{\LARGE \bf Performance evaluation of a foot-controlled human-robot interface}
\author{Yanpei~Huang, Etienne~Burdet, Lin~Cao, Phuoc~Thien~Phan, Anthony~Meng~Huat~Tiong, Pai~Zheng, and~Soo~Jay~Phee 
\thanks{All authors are with the School of Mechanical and Areospace Engineering, Nanyang Technological University, Singapore. Etienne Burdet is also with the Department of Bioengineering, Imperial College of Science Technology and Medicine, London, UK.}}
\maketitle

\begin{abstract}
Robotic minimally invasive interventions typically require using more than two instruments. We thus developed a foot pedal interface which allows the user to control a robotic arm (simultaneously to working with the hands) with four degrees of freedom in continuous directions and speeds. This paper evaluates and compares the performances of ten naive operators in using this new pedal interface and a traditional button interface in completing tasks. These tasks are geometrically complex path-following tasks similar to those in laparoscopic training, and the traditional button interface allows axis-by-axis control with constant speeds. Precision, time, and smoothness of the subjects' control movements for these tasks are analysed. The results demonstrate that the pedal interface can be used to control a robot for complex motion tasks. The subjects kept the average error rate at a low level of around 2.6\% with both interfaces, but the pedal interface resulted in about 30\% faster operation speed and 60\% smoother movement, which indicates improved efficiency and user experience as compared with the button interface. The results of a questionnaire show that the operators found that controlling the robot with the pedal interface was more intuitive, comfortable, and less tiring than using the button interface.
\end{abstract}
 
 \section{Introduction}
Robotic surgery often involves three or four robotic arms, e.g., two for interventions and one for camera \cite{EndoMaster2017}. It is desirable that the surgeon can control all the arms simultaneously without any additional assistants. This may improve the efficiency and safety of surgical operation by avoiding communication errors with (human) assistants. Therefore, interfaces using speech or the head, foot or finger movements \cite{Buess2000, Polet2008, Endex1993} have been used to position the laparoscopic camera from time to time. The studies \cite{Abdi2015, Abdi2016} have demonstrated how a ``third arm'' can be controlled by a foot to work together with the hands in continuous motions.

The most common foot interfaces in surgical applications use switches or buttons to control movement in one degree of freedom, e.g., to activate the bipolar forceps \cite{Hangingswitch2013}, or interact with the image in radio-logical interventions \cite{mci/Hatscher2018}. They can also be used to activate multi-functions, such as the phaco foot pedal controlling the irrigation, aspiration and ultrasonic power delivery in phacoemulsification \cite{Phacopedal2014}, or as a console to control a microscope's focus, zoom, field of view centring and light \cite{Thorlakson1998}.
 
Foot interfaces to control the movements of the laparoscopic camera \cite{Sackier1994, ViKY2010, RoboLens2011} traditionally consist of multi-directional switches placed on a planar platform, where each switch moves the robot in one direction with constant speed. Such interfaces are relatively easy to use for simple tasks. However, the operation will become difficult and less efficient for spatial movements of the robot when frequent or continuous direction and force adjustments are required. The number of buttons will grow with the number of directions to be controlled, increasing the complexity of operation and the risk of mistakes. The operator should be able to plan the movement in discrete single Cartesian directions, identify correct buttons without looking at the buttons, and then carry out a suitable pressing sequence and time. These steps may cause fatigue with float-in-air foot gestures and mental effort to select a correct buttons sequence. Therefore, we have developed an alternative ``pedal interface'' to control a robot, that can provide continuous control in all directions and of speed magnitude, and may thus address these issues with interfaces made of simple switches \cite{2018Huang}.
 
 Hand control to teleoperate a robotic arm for surgery has been extensively studied, e.g. \cite{DaVinci2004, Tobergte2011, Raven2}, where the hands' position is intuitively mapped to the position of the robot. However, much less work has been carried out on teleoperation control by foot. In this study, ten participants performed a teleoperation task using the dominant foot to control a slave robot in three translations and one rotation. The experiment involved a path following task which requires accuracy and dexterity in four degrees of freedom (DOFs). We compare the performance obtained from our pedal interface with those using a traditional button interface. Performances are analyzed in terms of the operation error rate, task completion time and motion smoothness.
 
 The rest of paper is organized as follows. Section II introduces the teleoperation system as well as the pedal and button interfaces to be compared. Section III then describes the protocol of the user study. Section IV presents and discusses the experimental results. Section V concludes the paper.

\section{Teleoperation system}
\begin{figure*}[!t]
\centering
\includegraphics[width=0.85\textwidth]{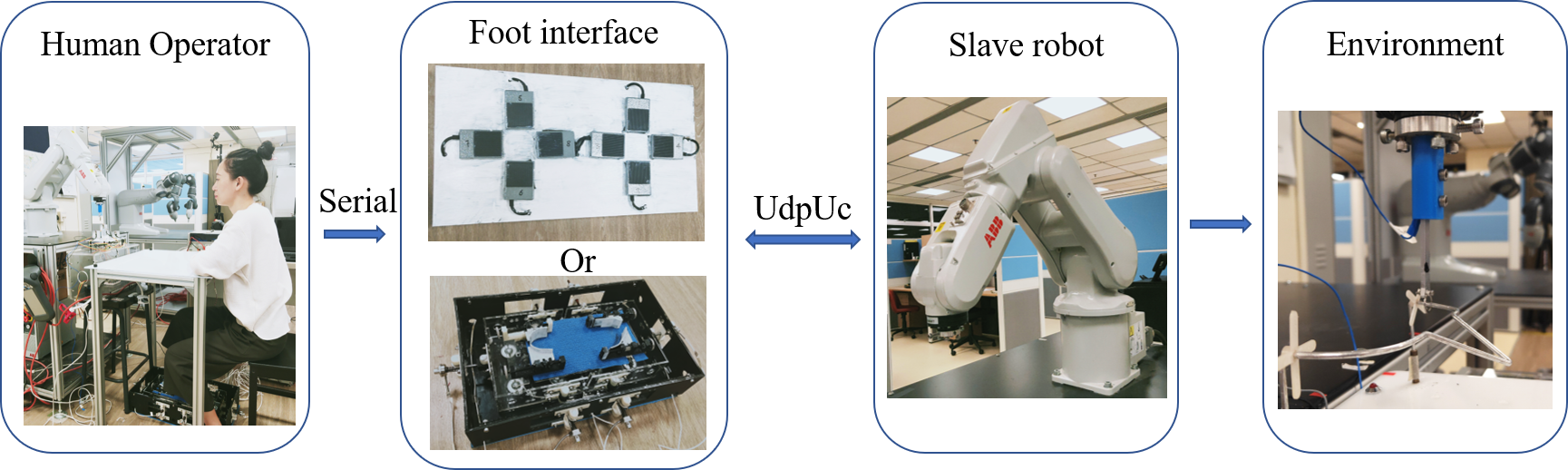}
\caption{Teleoperation system framework.} 
\label{f:teleoperation system}
\end{figure*} 

The teleoperation system used in our skillful operation experiment, shown in Fig. \ref{f:teleoperation system}, includes the master and slave devices. The human operator moves the foot in the pedal or button foot interface (Section \ref{s:master_interface}), the continuous or discrete foot information is then transmitted to the computer through serial communication. The interface's output is then mapped to four-DOF robot velocity control commands (Section \ref{s:slave}) through mapping models of foot interfaces (Section \ref{s:mapping}).  The corresponding robot configuration is then sent to the robot controller through the User Datagram Protocol Unicast Communication (UdpUc), in which the computer acts as the server and the robot controller is the client.

\subsection{Interfaces} \label{s:master_interface}
\begin{figure}[!t]
\centering
\includegraphics[width=0.5\textwidth]{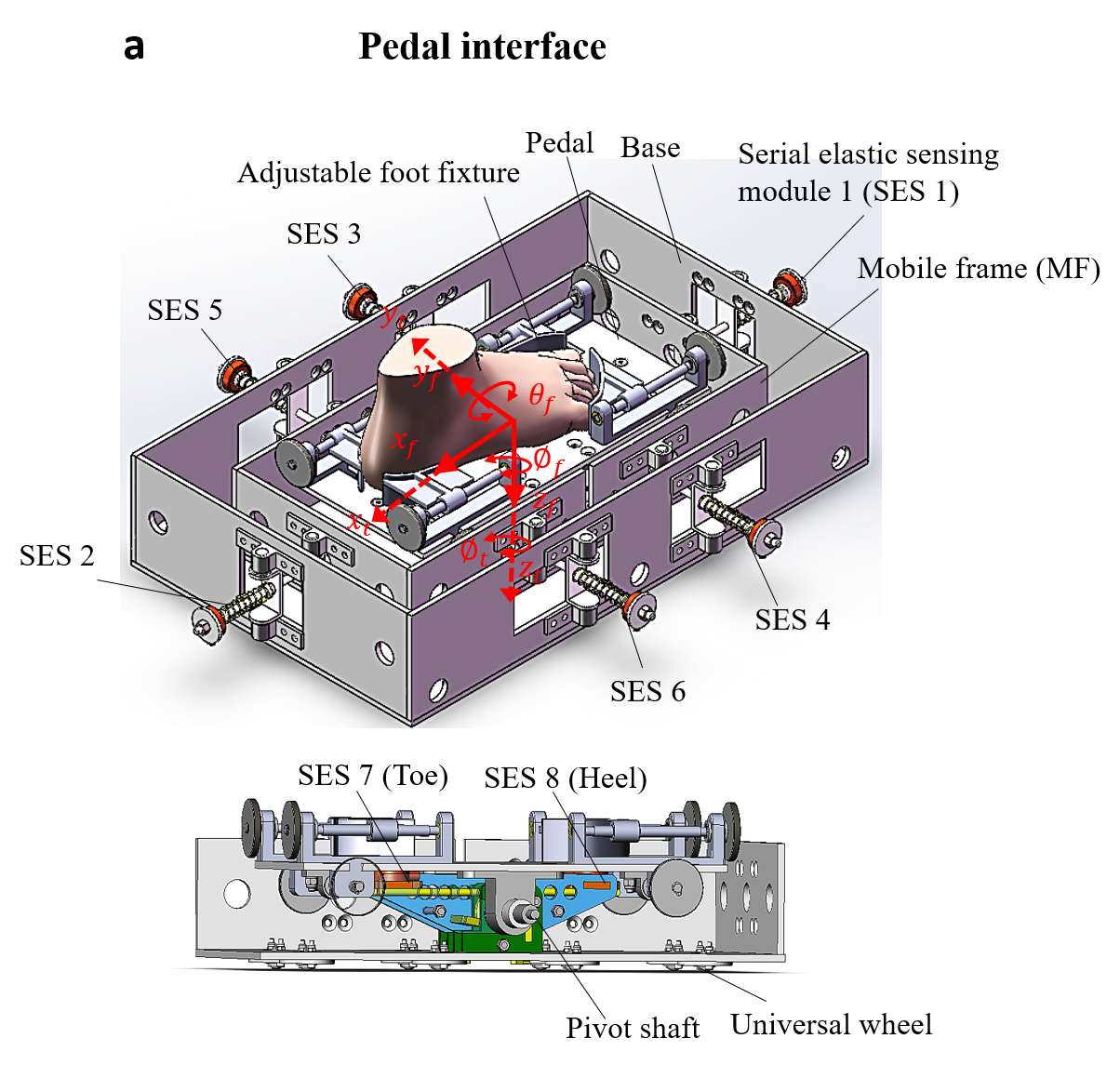}
\includegraphics[width=0.4\textwidth]{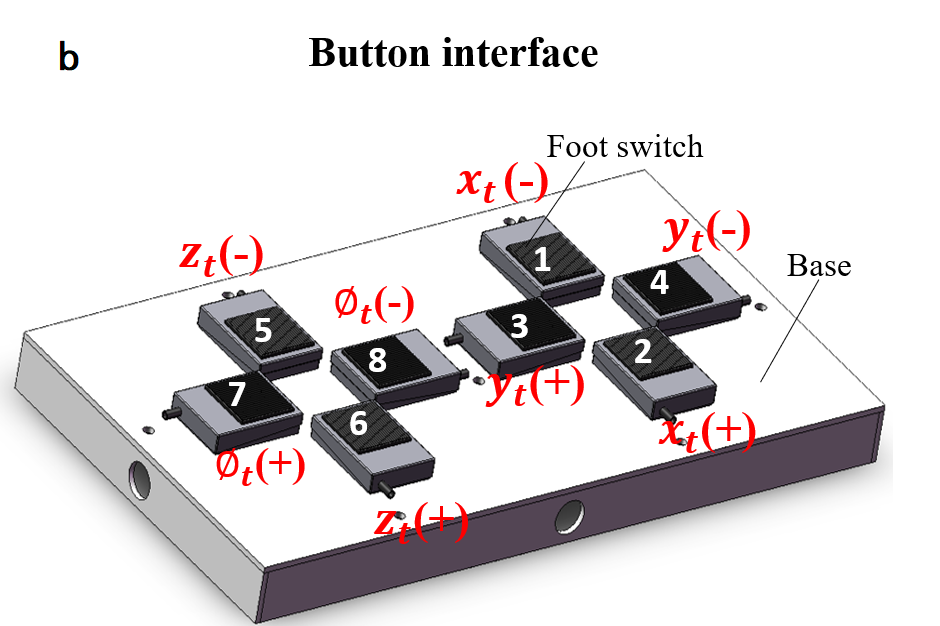}
\includegraphics[width=0.45\textwidth]{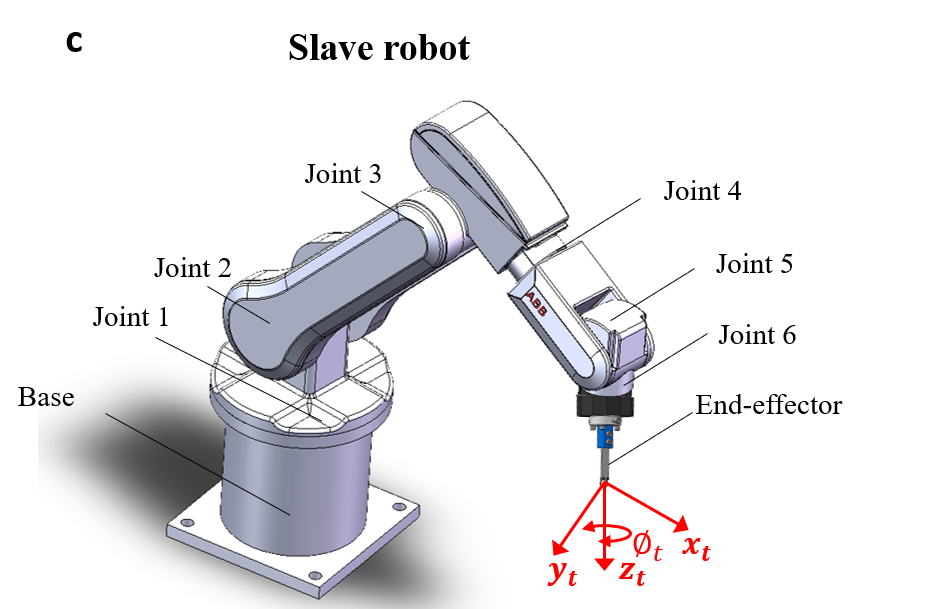}
\caption{Interfaces to control the robot; (a) pedal interface in perspective top and open side views, (b) button interface coding robot movement directions, and (c) slave robot.} 
\label{f:pedal}
\end{figure}

\subsubsection{Pedal interface}
The pedal interface (\(568 \times 372 \times 102mm\)) shown in Fig. \ref{f:pedal}a is designed to acquire the continuous movements of the foot which interacts with the pedal. The mobile coordinate of the pedal plate is represented as  \(x_f,y_f,z_f\). The foot movements in translations of \(x_{f},y_{f}\) and rotations around \(y_{f}\) and \(z_{f}\), represented as \(\theta_f\) and \(\phi_f\) and any possible combinations of them can be collected. The pedal interface has a parallel-serial structure consisting of a base, a mobile frame (MF), a pedal plate with adjustable foot fixture and eight serial elastic sensing modules (SES). Each SES is composed of a serial connected elastic element or spring and a load cell, which senses the applied force from the deformed elastic element thus the position. The measured forces can be used to calculate the deformation of the springs and thus the position of the foot. Once the elastic element reaches the motion limit, isometric force can further be recorded. To reduce the friction of dragging the foot pedal, eight universal wheels are mounted at the bottom of the MF. 

As natural movements in desired directions may not correspond exactly to the interface axes, and may further exhibit subject-specific patterns, a data-driven calibration procedure (see Section \ref{s:experiment}) was developed to capture subject specific motion patterns. A \(4\times 8\) transformation matrix is derived from the calibration data using the independent component analysis (ICA), which directly maps the eight force sensors signals to four-DOF control commands \cite{2018Huang}. This transformation defines motion axes based on the individual operator's motion characteristics.  

\subsubsection{Button interface}
The button interface ($652 \times 356 \times 50 mm$) shown in Fig. \ref{f:pedal}b was designed similar to existing commercial foot interfaces \cite{Sackier1994, ViKY2010, RoboLens2011} but with additional DOF control. The four DOFs of the robot are controlled by eight foot switches ($90 \times 66 \times 24 mm$) with \(55 \times 55 mm\) pressing area. They are located in two button areas, with each consists of four foot switches. Pressing one foot switch can activate the movement in the associated single Cartesian direction at a constant speed. All the foot switches are activated by the same foot gesture of lifting up the foot and pressing down. Note that the operator may skillfully press two buttons simultaneously with one foot in order to command combined movement in two directions. The button interface is advantageous to control motion in single directions without much direction deviation. However, more complicated tasks requiring movement combining multiple directions and frequent redirection, which lead to long operation time, fatigue to the operator and yield jerky zigzag trajectory. 

\subsection{Slave robot} \label{s:slave}
The robotic arm to be controlled is a six-DOF manipulator IRB120 (ABB Robotics, Switzerland) with serial structure described Fig. \ref{f:pedal}c. The slave robot is controlled using the external guided motion (EGM) mode \cite{ABB_EGM}, which bypasses the path planning procedure and enables a 30Hz response to commands from the foot interface. A low-pass filter with 10Hz cut-off frequency was used between the EGM controller and the robot motion controller.

\subsection{Mapping to the robot coordinates}\label{s:mapping}
For the pedal interface, the coordinates of the foot \(x_f,y_f,\theta_f\) are matching the frame \(x_t,y_t,\theta_t\) of the robot (Fig. \ref{f:pedal}a). The \(\phi_f\) maps to \(z_t\) with toe down rotation correspond to \(z_t\) positive and toe up rotation control \(z_t\) negative. The slave robot is operated in speed control without work-space limitation. The robot velocity is proportional to the force exerted by the foot. A dead zone is used for each DOF control command to prevent the robot from moving with small involuntary foot movement. The maximum speed is limited to a target value based on different applications. 

For the button interface, the foot switches are arranged to match the spatial azimuth relationship (Fig. \ref{f:pedal}b). The left four switches correspondingly control the translation and rotation of \(z_{t}\) axis. Button 5 and 6 control \(z_{t}\) negative and positive, respectively; button 7 and 8 map to the anti-clockwise and clockwise rotation around \(z_{t}\) axis. The right four switches correspondingly controls the translations of horizontal plane of the robot, i.e., button 1 and 2 activate the movements in \(x_{t}\) negative and positive directions; button 3 and 4 control positive and negative directions along \(y_{t}\). The eight button states in binary form are sending to the master computer and each pressing will command the robot to move in the corresponding direction with a constant speed.

\section{Experiment}\label{s:experiment}
The experiment was approved by the Institutional Review Board (IRB) of Nanyang Technological University (IRB-2018-05-051). Ten participants (five males and five females) with average age of $28.7\pm2.2$ years, without motor impairment, were recruited for the experiment. These participants were all right-footed according to the ball-kick dominant leg test \cite{2017kickball}, and none of them regularly used any foot based gesture systems. They received information about the purpose and protocol and signed the informed consent form before the experiment.

\subsection{Setup}
\begin{figure}[!t]
\centering
\includegraphics[width=0.45\textwidth]{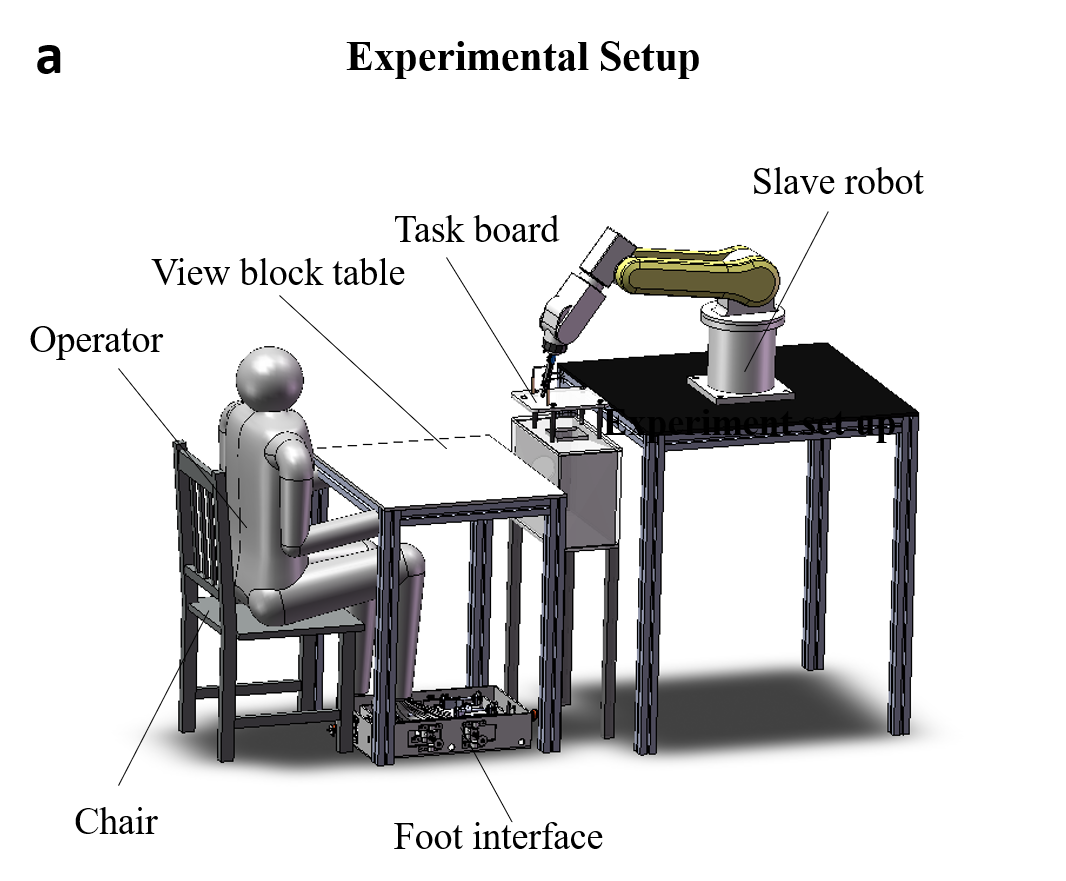}
\includegraphics[width=0.51\textwidth]{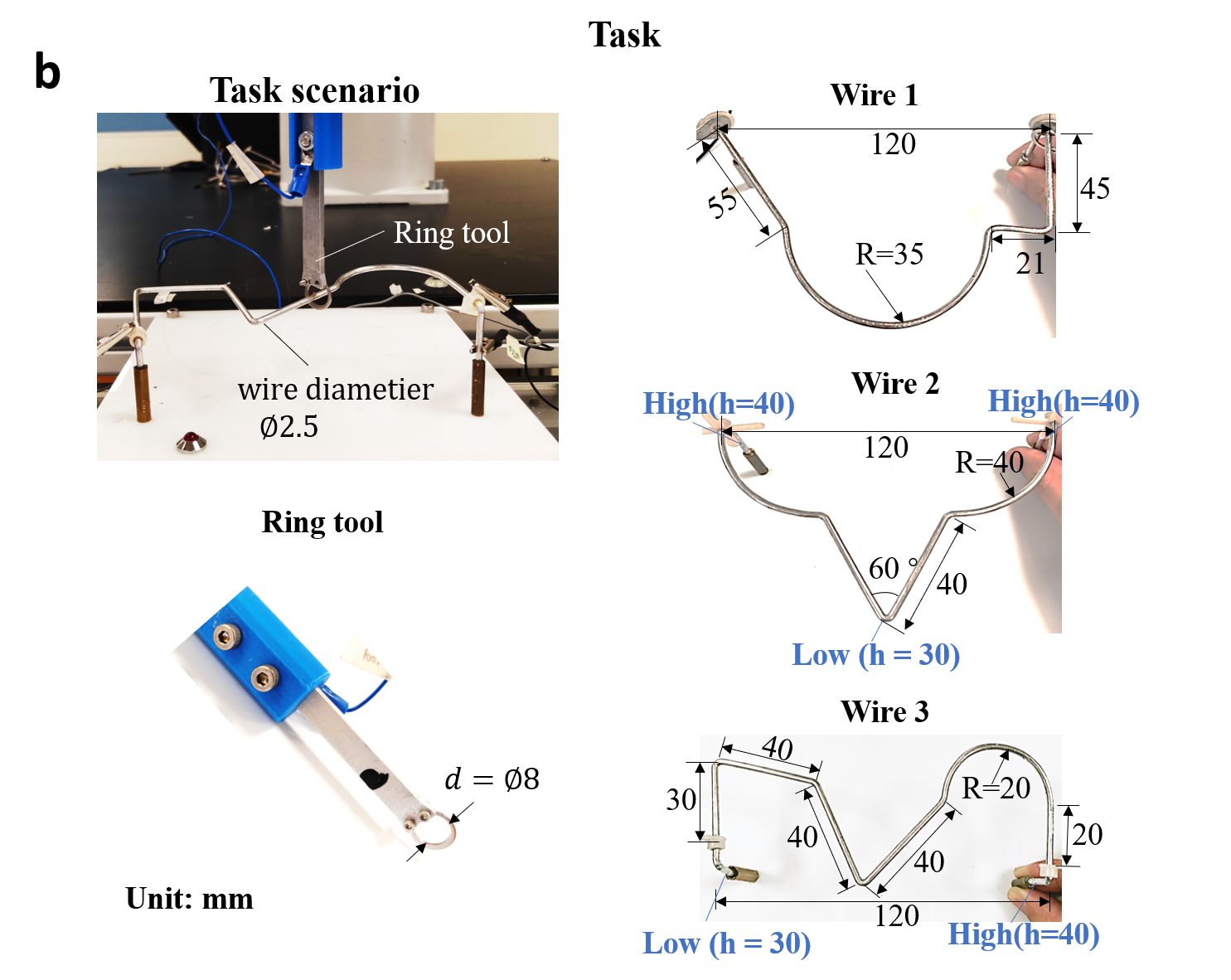}
\includegraphics[width=0.5\textwidth]{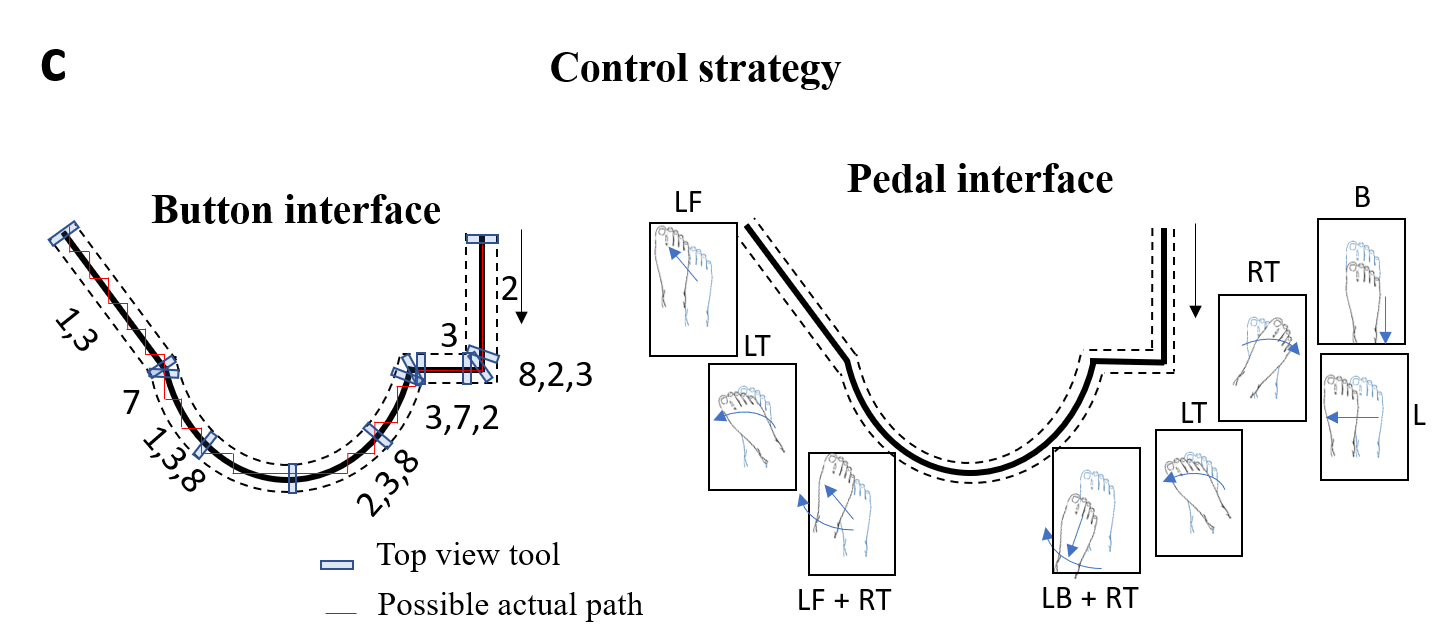}
\caption{ Experiment overview. (a) Experimental setup. (b) Task scenario (left top), enlarged views of robot end-effector tool (left bottom) and three training and testing wires (right). (c) Possible control strategy for the button interface (The numbers on the left panel represent corresponding buttons) and the pedal interface (B: backward, L:left, RT: right rotation, LT: left rotation, LB: left and backward, LF: left and forward).} 
\label{f:task}
\end{figure}

Either foot interface was placed on the ground in the front of the subject, who comfortably sit on a fixed chair and controlled the end-effector (using dominant foot) by directly looking at the end-effector (Fig. \ref{f:task}a). A block view table (\(660 \times 580 \times 800mm\)) placed between the foot interface and the participant to prevent the operator from watching the foot or the interface during operation. For each interface, the subject was given five minutes to get familiar with its operation before the test trial. The default mappings between each of the foot interfaces and the slave robot are depicted in Section \ref{s:mapping}.

\subsection{Calibration} 
For the pedal interface, the calibration data were collected through foot movements in eight directions: forward (F), backward (B), left (L), right (R), toe up rotation (TU), toe down rotation (TD), left torsion (LT), and right torsion (RT). Starting at the home position, the pedal is moved smoothly in each specified direction to the boundary of the workspace, the pedal is held for one second and the foot returns back to the home position \cite{2018Huang}. Three movements were conducted in each direction. This procedure was carried out continuously until all the \(3\times8 = 24\) centre-out and back movements were completed. This calibration data were used to identify the independent component analysis (ICA) mapping model, then the participant given two minutes to test the mapping model with robot movement.

\subsection{Protocol}
The task designed is similar with the laparoscopic training task\cite{Schreuder2011}, but conducted in a teleoperation system and using foot control. The participant was asked to move the robot end-effector along each of the three paths as fast and accurately as possible without hitting the wire. The test scenario, robot end-effector tool and testing wires are shown in Fig. \ref{f:task}b. A small ring with 8mm inner diameter is connected to the robot end-effector. The ring needs to be guided along an aluminum wire (2.5 mm diameter) with specially designed shapes. Three wires with different shapes were used. The three wires define different paths including 1) single Cartesian path along $x_t$, $y_t$, 2) two-DOF diagonal path in $x_t-y_t$ plane, 3) two-DOF translation and one-DOF rotation circle path, 4) the above paths and combined translation in $z_t$ and 5) turning points with angles of equal, less or larger than $90^o$. The wire 1 is a 2D horizontal path. The second and third paths are 3D routes requiring translation in $z_t$ with 10mm. Experiment with the first two paths is considered as training and conducted on a first day while the third path is used on a second day to test the acquired skill. Half of the participants start (on both days) using the pedal interface and then the button interface while the other subjects start with the button interface. 

The starting position of the ring tool is at one side of the wire. One trial is completed when the tool is moved from one side of the path to another side. The odd-numbered trials start from left to right, and the others start form right to left. There is a 10-second short break between consecutive trials. After finishing 10 trials at wire 1 with one interface, the subject relaxes for a 2 minutes break, after which wire 2 is used. After finishing practicing with these two wires with one interface, and a 5 minutes break, the same protocol is followed with the other foot interface. Test on wire 3 is carried out similarly on the second day. 

The maximum speed for the robot end-effector was set to $6mm/s$ in translation and $10^o/s$ in rotation when controlling with the pedal interface, which correspond to the constant speed used with the button interface. These limits were set in preliminary trials carried out by the experimenter to offer comfortable control.

\subsection{Performance measures}
A relatively strict accuracy constrain is selected for foot control of within $\pm 2.75mm$ in $x_t, y_t$ and $z_t$, which is similar to the typical accuracy of of 1 to 5mm in hand teleoperation systems \cite{2004hand}. The expected control strategy for both interfaces are shown in Fig. \ref{f:task}c with path 1 in top view as an example. The dash lines shows the the maximum allowable deviation when the ring tool's cross-section is perpendicular to the path and concentric with the wire circle cross-section. The participants' performance with foot control is assessed through the error rate, completion time, smoothness and the subjective questionnaire.

\paragraph{Tracking error rate}
The operator needs to dynamically adjust the ring via foot control while avoiding touching the wire. When the wire is touched, the buzzer is on which is recorded at 20Hz. The error rate is the percentage of touching time divided by the completion time in the same trial.
 
\paragraph{Completion time}
The completion time is the most direct measure index of performance in teleoperation control. The time is recorded when the robot starts to move (i.e. when quitting the start metal plate) until the ring tool touches the end metal block. The task is carried out back and forth, where one trial corresponds to either the forth or back movement.

\paragraph{Smoothness}
Motion smoothness at the slave robot can be used to reflect the foot motion control performance. How smooth or jerky the movements of the slave robot is quantified using the absolute value of the spectral arc length smoothness metric \cite{Balasubramanian2012}.

\begin{table*}
\centering
\caption{Subjective assessment questionnaire}
\begin{tabular}{|p{6.3cm}|c|c|c|c|c|c|c|c|}
\hline
\multicolumn{2}{|c|}{Statement} & \multicolumn {5}{|c|}{Score} &\\\hline
1. The mapping between foot and robot movements is & not intuitive &1&2&3&4&5& very intuitive\\\hline
2. Accurately following the path was & difficult &1&2&3&4&5& easy\\\hline
3. The mental effort required for operation was & low &1&2&3&4&5& high\\\hline
4. Foot fatigue was & none &1&2&3&4&5& very high\\\hline
5. Operation speed was & too low &1&2&3&4&5& too fast\\\hline
6. General comfort was & very uncomfortable &1&2&3&4&5& very comfortable\\\hline
7. Overall the input device was & difficult &1&2&3&4&5& easy to use\\\hline
8. Movement during the operation was & rough &1&2&3&4&5& smooth\\\hline
\multicolumn{8}{|l|}{9. In general, which interface do you prefer? Why? Please specify your reason(s)}\\
\hline
\end{tabular}
\label{t:questionnaire}
\end{table*}

\paragraph{Questionnaire}
At the end of the task, participates were given a questionnaire (Table \ref{t:questionnaire}) with nine questions to assess the use of the two interfaces to control the robot. The first eight statements are rated on a discrete five-point Likert scale.

\section{Results}
\begin{figure*}[!t]
\centering
\includegraphics[width=0.95\textwidth]{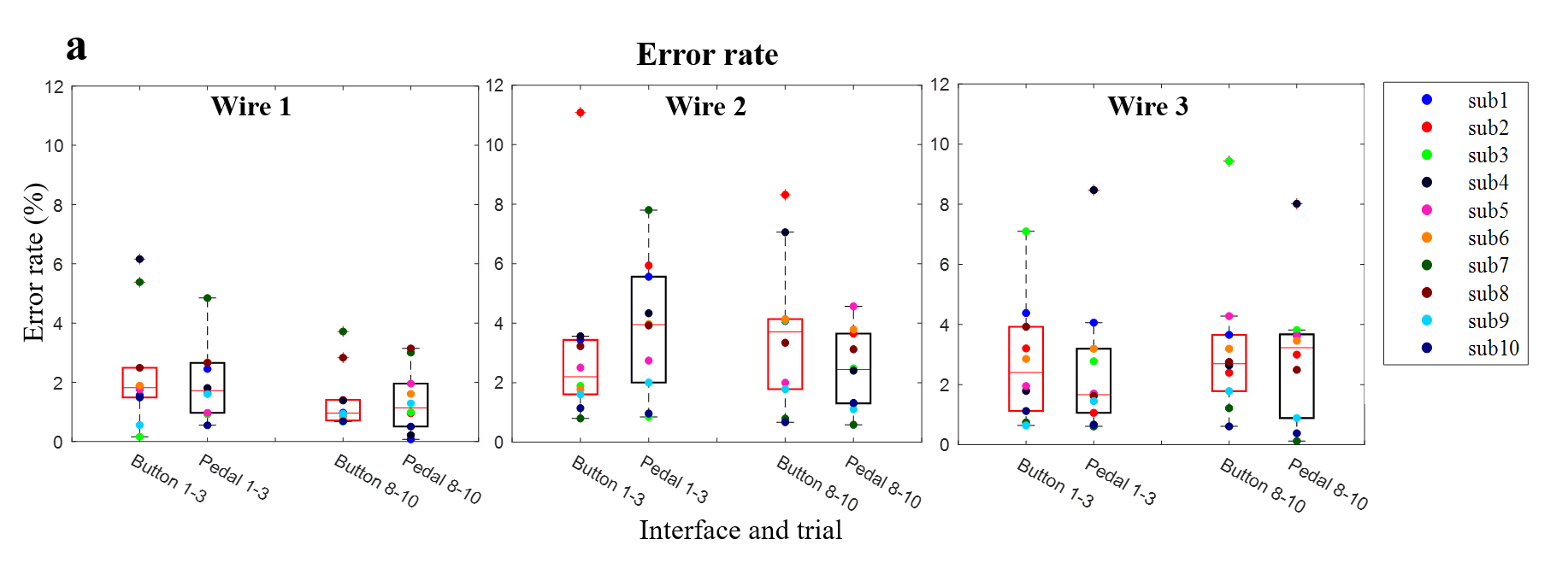}
\includegraphics[width=0.95\textwidth]{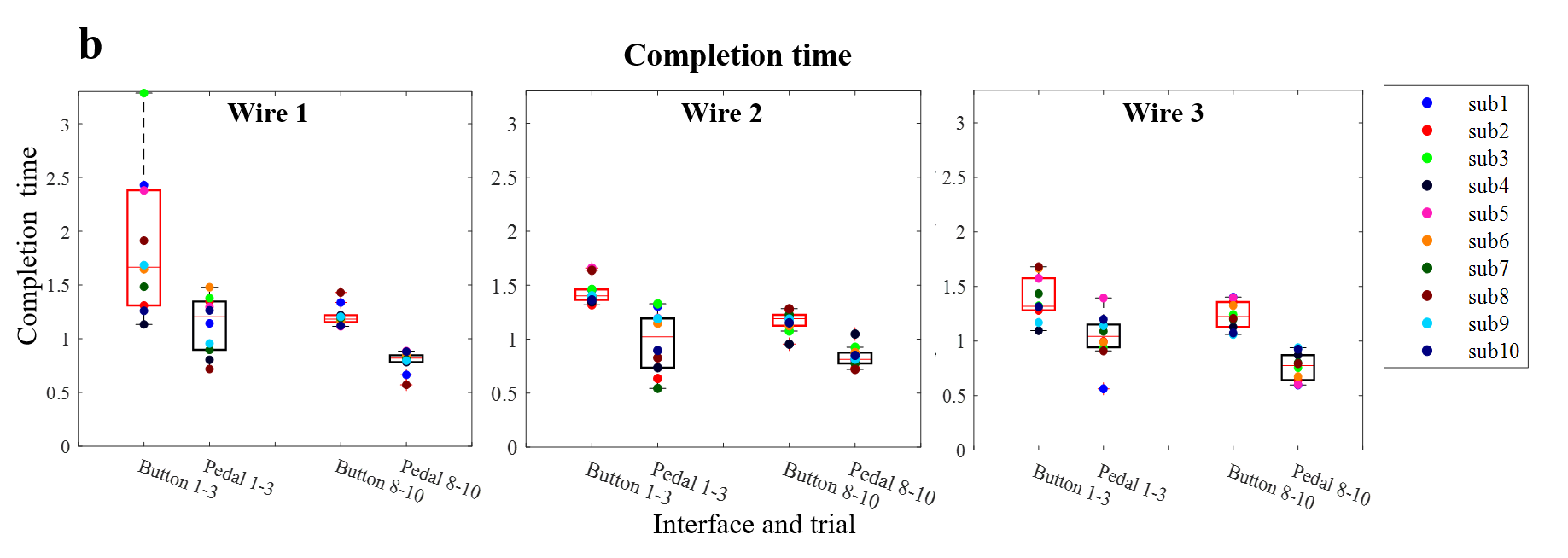}
\includegraphics[width=0.95\textwidth]{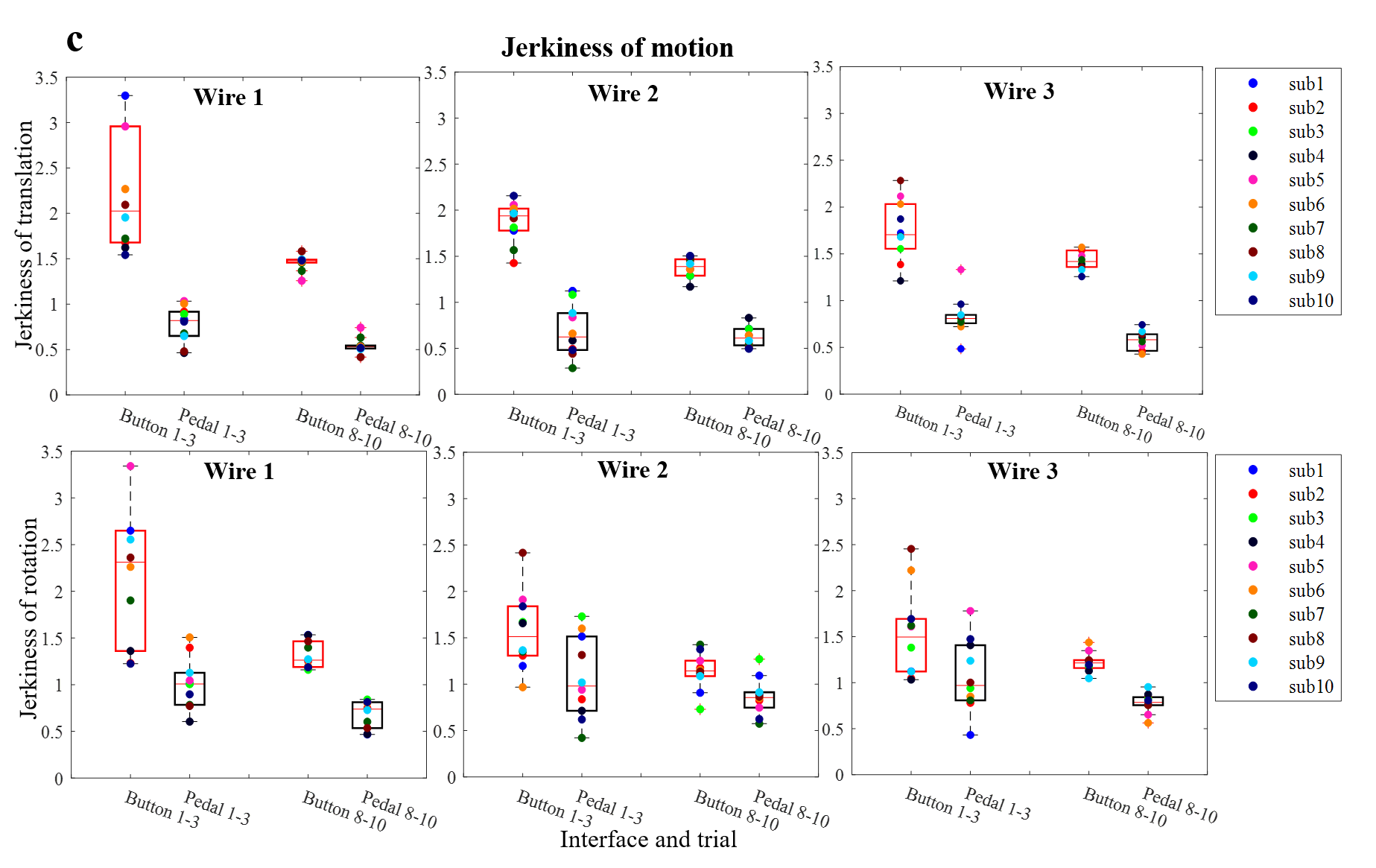}
\caption{Analysis of performance in the three tasks; Average (a) error rate (b) completion time and (c) motion jerkiness of first and last three trials for the button and pedal interfaces of wire 1, wire 2 and wire 3.} 
\label{f:operation_performance}
\end{figure*}

\subsection{Behavioural analysis}
 No significant difference was found on any of the metrics of error rate, completion time and motion smoothness between participants starting with the pedal vs. button interface, moving from left to right vs. right to left, thus the data of these sub-populations were treated together. The results were also gender independent. The results in Fig. \ref{f:operation_performance} show the average metrics values over the first and last three trials.

The result of foot motion control accuracy is shown in Fig. \ref{f:operation_performance}a. The effect of two foot interfaces is checked using T-test with $p = 0.05$ confidence interval. No difference was found in the error rate obtained with the two foot interfaces ($p > 0.13$). The average error rate was small, generally 1-4\%. We interpret this as that the subjects kept the error at an admissible low rate with both interfaces.

One can find in Fig. \ref{f:operation_performance}b that the pedal interface enables to reduce completion time ($p < 0.006$) by about 30\% in all three paths relative to the button interface. Compared to the button interface operation, the pedal interface allows the operator to carry out movements combining multiple DOFs and directions, without having to decompose the movement along separate DOF as with the button interface. Furthermore, each move with the button interface requires identifying the correct button for each phase/direction and pressing it, which is time-consuming.

Fig. \ref{f:operation_performance}c analyzes the average jerk at the slave robot in translation (Fig. \ref{f:operation_performance}c, top row) and rotation (Fig. \ref{f:operation_performance}c, bottom row) separately. We see in this figure that the robot has jerky movements when controlled by the button interface. The operator has to conduct many futile movements from one button to another, but only the pressing action yields the robot motion commands. In contrast, the spectral smoothness index of translation increased (\( p < 0.014 \)) by 63\%, 59\% and 56\% using the pedal interface allowing continuous direction changes relative to the button interface for the three tracking paths. The rotation of the robot also become smoother (\( p < 0.045\) with exception of the second path with (\( p = 0.060\), by 48\%, 22\%, 30\% for three paths respectively when using the pedal interface than the button interface.

Learning can be observed through the completion time and jerk. The completion time for last three trials reduced by about 28\% for both button and pedal interfaces in the first path compared to the initial three trials. The learning rate slowed down on the second path, perhaps due to the learning experience along the first path. The third path is the testing path and conducted at the second day, where the results suggest some re-learning. The pedal interface yielded a faster learning rate with 24\% reduction on time from three first to last trials, the value for the button interface was 9\%. The metric of jerkiness on translation exhibit a similar learning effect.

\subsection{Subjective assessment}
\begin{figure*}[!t]
\centering
\includegraphics[width=0.77\textwidth]{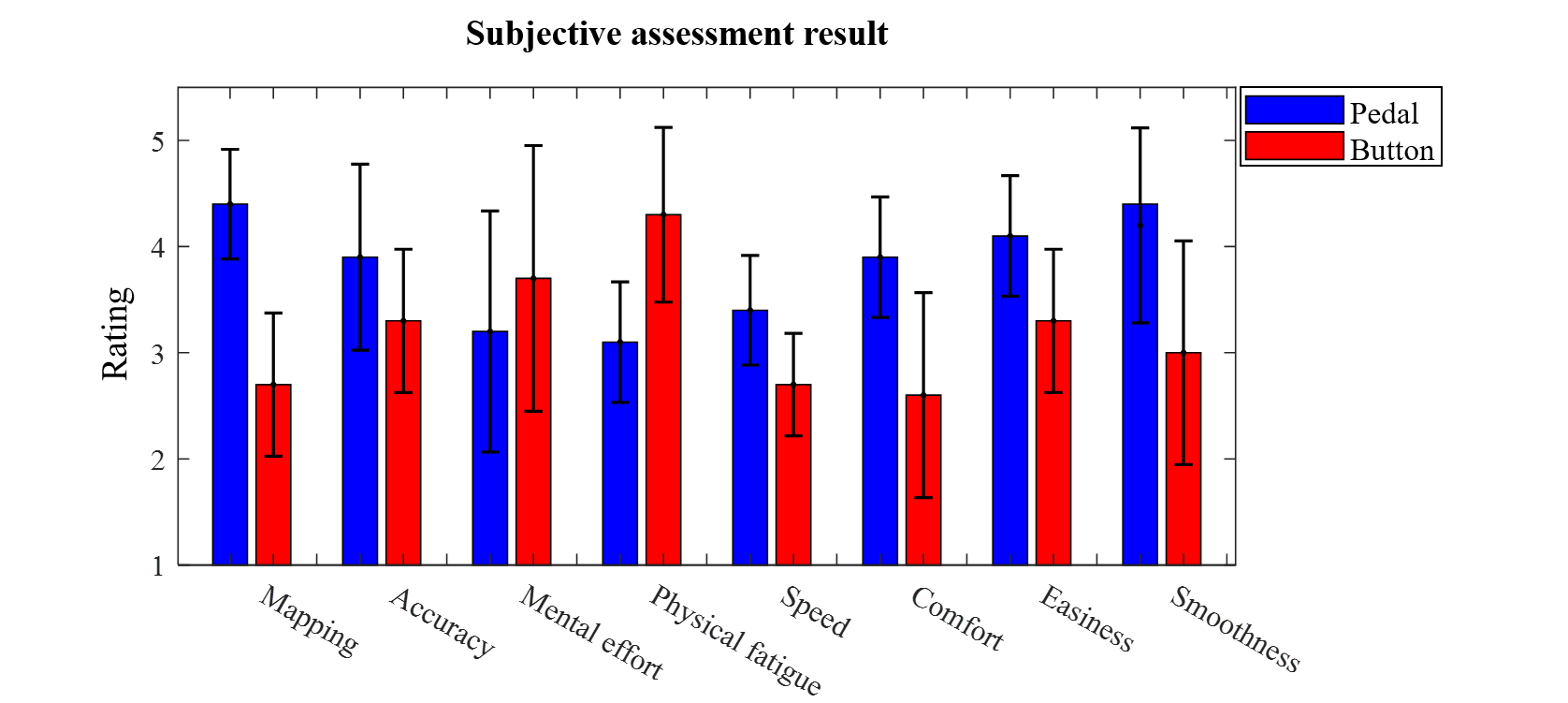}
\caption{Subjective assessment result of the pedal and the button foot interfaces.}
\label{f:questionnaire}
\end{figure*}

Fig. \ref{f:questionnaire} illustrates the average responses and the standard deviation of ten participants to the questionnaires of Table \ref{t:questionnaire} about the robot control with the two interfaces. An Analysis of variance (ANOVA) revealed that the two different foot control strategies had a significant effect on the rating of \emph{intuitiveness} ($F_{0.05,1,18} = 40.02, p< 0.00001$). Participants found the mapping is more intuitive using continuous foot movements in all directions (the mean responses are 4.4 for the pedal interface, 2.7 for button interface). The ANOVA also revealed that the foot control strategies had a significant effect on the \emph{physical fatigue} ($F_{0.05,1,18} =14.4, p = 0.0013$). The button interface (\(4.30\pm 0.82\)) was felt to require larger physical effort than the pedal interface (\(3.10 \pm 0.56\)). Significant effects were also found for the criteria of \emph{comfort} ($F_{0.05,1,18} =13.46, p = 0.0018$), \emph{ease of use} ($F_{0.05,1,18} =8.23, p =0.0102$), and \emph{smoothness} ($F_{0.05,1,18} =7.36, p =0.0142$), participants generally found the pedal interface had superior performance in those aspects. It is interesting that participants felt the robot moved at faster speed (ANOVA: $F_{0.05,1,18} = 9.8, p =0.0058$) using the pedal interface, although speed was smaller or equal to the constant speed used with the button interface. No significant effects was found for the \emph{accuracy} ($F_{0.05,1,18} =2.95, p =0.1033$), which corresponded to the objective error rate results. Also \emph{mental fatigue} ($F_{0.05,1,18} = 0.88, p =0.3618$) was not deemed larger with the button interface. The pedal interface needs continuous focus on the movement direction, while the button interface requires the subject pay more attention on the motion-button mapping and motion planning. In fact, 9/10 of the participants clearly preferred the pedal interface due to the more efficient operation and reduced fatigue. 

\section{Conclusion}
This paper compared different foot control strategies to operate a robot, using a new foot pedal interface and a traditional foot button interface. Ten participants were recruited to control the robot's end-effector to follow three different desired trajectories. A clear learning trend of operating both interfaces was observed from the performance of the participants. It was found that while the two interfaces enabled similar movement accuracy, they exhibited differences in the operation time, fatigue, and smoothness of trajectories. The button interface resulted in more physical fatigue, longer operation time, and less smooth trajectories as compared to the foot pedal interface. 

These differences can be explained by the different motion control principles offered by the two interfaces. For the button interface, movements in different DOFs are separately controlled by the associated buttons, which results in discrete movements, time-consuming motion adjustment, and less smooth trajectories. In addition, participants needed to frequently lift/drop the foot to release/press the buttons for multi-DOF motion control, which prolonged motion adjustment and significantly increased physical fatigue.  In contrast, for the four-DOF foot pedal interface, the natural continuous movement of the foot in multi-DOF is directly mapped to the motion of the robotic arm, leading to more intuitive and efficient operation and smoother trajectories. 

Following the presented preliminary results, further studies of foot-controlled interfaces will be conducted, e.g., the motion/force control capabilities of the foot, more accurate control with robotic assistance, as well as the design and assessment criteria for foot interfaces in teleoperation surgical systems. In addition, the presented work used an industrial robot instead of a surgical robot. Although the industrial robot can demonstrate the performance of the interfaces in accomplishing complex motion tasks similar to those in laparoscopic training, performance of the pedal interface in a more realistic surgical scenario using a surgical robot will be investigated.

\section*{Acknowledgment}
We thank the subjects for taking time to carry out the experiment. This work was funded in part by the Singapore National Research Foundation through the NRF Investigatorship Award (NRF-NRFI 2016-07), and by the EU-H2020 ICT-644727 COGIMON grant.

\addtolength{\textheight}{-12cm} 

\bibliographystyle{IEEEtran}
\bibliography{IEEEabrv,reference}

\end{document}